\documentclass[conference]{IEEEtran}
\IEEEoverridecommandlockouts
\usepackage{lineno,hyperref}
\usepackage{cite}
\usepackage{amsmath,amssymb,amsfonts}
\usepackage{algpseudocode}
\usepackage[linesnumbered,ruled,vlined]{algorithm2e}
\usepackage{adjustbox}
\usepackage{graphicx}
\usepackage{subfig}
\usepackage{threeparttable}
\usepackage{textcomp}
\usepackage{xcolor}
\usepackage{multicol}
\usepackage{multirow}
\usepackage{tabularx}
\usepackage{parskip} 

\def\BibTeX{{\rm B\kern-.05em{\sc i\kern-.025em b}\kern-.08em
    T\kern-.1667em\lower.7ex\hbox{E}\kern-.125emX}}
\begin{document}

\title{A multimodal deep learning framework for scalable content based visual media retrieval}

\author{\IEEEauthorblockN{Ambareesh Ravi}
\IEEEauthorblockA{\textit{Dept. of Electrical and Computer Engineering} \\
\textit{University of Waterloo}\\
Waterloo, Canada \\
ambareesh.ravi@uwaterloo.ca}
\and
\IEEEauthorblockN{Amith Nandakumar}
\IEEEauthorblockA{\textit{Dept. of Electrical and Computer Engineering} \\
\textit{University of Waterloo}\\
Waterloo, Canada \\
a4nandakumar@uwaterloo.ca}
}
\maketitle

\begin{abstract}
We propose a novel, efficient, modular and scalable framework for content based visual media retrieval systems by leveraging the power of Deep Learning which is flexible to work both for images and videos conjointly and we also introduce an efficient comparison and filtering metric for retrieval. We put forward our findings from critical performance tests comparing our method to the predominant conventional approach to demonstrate the feasibility and efficiency of the proposed solution with best practices, possible improvements that may further augment ability of retrieval architectures.
\end{abstract}


\section{Introduction}
Recent developments in the field of digital media have
led to an abundance of visual data around us. Organizing, managing and gaining insights out of the visual data has become more difficult owing to the sheer volume of the data. Hence, tasks like visual search, reverse search, retrieval, captioning, indexing have become more crucial over the past years. The need for an automated and efficient approach to index, categorize and organize visual media with little to no human intervention is growing in multiple fields like medicine, satellite imagery, remote sensing, forensics, historical search, media recommendations, Video on Demand, online shopping, security and digital libraries etc.

The advent and recent popularity of content based retrieval techniques are due to the short-comings of text-query based retrieval methods. Query based techniques need comprehensive metadata tagging by humans involving a lot of effort in terms of annotations and tagging. They also function satisfactorily only when the input query matches the tags available in the data repository.  Whereas in content based methods, the information from the input image is directly correlated with that of the ones in the database.

Deep Learning has evolved over the years and has shown compelling performance at complex vision and natural language tasks than the conventional approaches \cite{singh2015content} with the potential to close the semantic gap between the low-level features such as color, shape, texture, edges, orientation etc., that is apparent to the learning algorithm and the high level context as perceived by humans through abstraction. In tasks like retrieval, the embedded features of the data are often compared and so the key emphasis is often on the ability of the model to produce a feature representation that can relate the similarities and discriminate the differences in the content.

The advances in the space of content-based video retrieval tasks has been less due to computational limitations, efficacy of algorithms and poor scalability due to the amount of data to be used, as videos contain temporal information in addition to spatial information. These limitations are taken into consideration and are mitigated considerably in our proposed framework which is scalable, flexible and efficient by nature. Accuracy and speed of the application is a critical element that draws the attention of the user and would be overlooked if one of the two is missing.

We propose a light, cogent and end-to-end content based retrieval architecture that works conjointly for both images and videos that uses Convolutional Neural Networks (CNNs) for spatial learning and Recurrent Neural Networks (RNNs) for learning the temporal relation that associates a sequence of frames into a video. We also provide substantiating evidence for the effectiveness of the proposed architecture in comparison with the traditional and prevalent approaches that are generally used for retrieval tasks. We also discuss the results of important experiments and the best practices that can further improve the performance, since the main focus of this work is to improve scalability of retrieval tasks. We concentrate primarily on the ability of our architecture to retrieve video as image retrieval is addressed adequately in other works.

\section{Related works}
In this section, we discuss a few important literature in the field of media retrieval. The three main areas of development are feature representation, ranking of retrieved results and faster comparison techniques.

Ladahke et al. \cite{solio2013review} provide a review of different retrieval tasks with their applications and early approaches which led to the development in the field as of now. The authors also illustrate the basic technique for CBIR tasks. In \cite{csurka2004visual}, Csurka et al., compare classifiers for visual categorization which is one of the field's earliest works after which improved methods emerged which led to the current popularity. After the advent of deep learning and the performance gains that come with it, several methods adopting deep learning techniques emerged . 

Both traditional methods and deep learning methods have their advantages and disadvantages. We focus on the facts as to why deep learning (DL) is the better option for content based retrieval tasks. DL is better at learning both local and global features where as traditional methods need separate modules to learn/identify local features such as shapes, texture, colour, edges, objects, orientation etc and they perform poorly in learning global features by association. DL models can be designed to be scale, spatial and colour invariant and can perform well in such cases subject to training on diverse data. Traditional image processing modules need separate pre-processing pertaining to their design and the features from different modules of traditional methods have to be aggregated to a fixed length and it is very difficult to maintain the correlation where as deep learning models readily output feature representations of fixed dimension.

In spite of several attractive features, deep learning models suffer from two major disadvantages – computational requirement and huge data necessity. But fortunately for retrieval tasks, the data is available in abundance and given that most of these tasks are hosted in cloud platforms which have high computational capabilities, deep learning is the most viable option given its performance.

In his thesis \cite{singh2015content}, the author A.V. Singh elaborates on the
differences between traditional and deep learning based
approaches and proves that the latter perform well over the traditional ones with their ability to
bridge the semantic gap considerably. \cite{wan2014deep} provides a
comprehensive study on deep learning for image retrieval
tasks by putting forward some compelling arguments on the
sub-par performance of traditional methods over learning
based methods. Traditional methods have several shortcomings in terms of complexity, inefficient operation and inability to represent a global generic solution etc which makes deep learning a favorable solution as they can be flexible, end-to-end and generalize over a wide range of data for various applications. In \cite{gordo2016deep}, A. Gordo et al., propose an approach to learn global representation for instance level retrieval using
deep learning. \cite{potapov2018semantic},\cite{arandjelovic2012three},\cite{gordo2017end}, \cite{babenko2015aggregating} provide the best practices for content based retrieval tasks which are
very helpful and insightful. Other favorable properties of deep learning are discussed in the next sections.

Since, the need for deep learning techniques was established in the previous sections, the popular works in the field are discussed in this section. There are several works on image retrieval and a very few notable ones on video retrieval. The main factor that decides the performance of any content based retrieval method is the feature representation or description of the input. There are many other works reflecting on the nature of feature representations and \cite{ismail2017survey} presents a detailed survey on the same.

The preliminary works in the field were based on visual cues such as shape, color, texture, edge and spatial features \cite{latif2019content} after which feature detection techniques such as SIFT, SURF, CENSURE and their variants were employed for better performance \cite{zhou2017collaborative}. Later, these paved way for the use of local features using sparse representations and bag of visual words (BoVW) \cite{latif2019content}. Performances of all these methods were surpassed by deep learning techniques. 
The seminal work \cite{krizhevsky2011using} uses autoencoders to abstract and learn the representations of images for content based retrieval tasks by mapping input images into concise 28-bit  binary codes. \cite{babenko2014neural} proposes to use the visual representations derived from the top layers of trained neural network to be used as visual encodings for retrieval. In \cite{zhu2016unsupervised}, the authors propose to use unsupervised deep learning methods to semantically hash images for retrieval by extracting vital information to improve the efficiency of visual hashing. \cite{wang2017survey}, \cite{rodrigues2020deep} present few other methods on hashing methods for image retrieval using deep learning. \cite{maji2020cbir} proposes to use ImageNet \cite{deng2009imagenet} pre-trained models as feature extractors for image retrieval tasks and compares the performance against other contemporary methods. \cite{wang2014learning} propose a deep ranking model to capture inter and intra class differences to improve the discerning ability of the model using triplet sampling. \cite{dubey2020decade} offers a comprehensive survey on the progression of deep learning based methods employed for CBIR over the decade with detailed analysis on the performance of the state of the art models.

Though there are numerous works on CBIR and text query-based video retrieval, there are only a few on content based video retrieval. The survey in \cite{patel2012content} elaborates on video retrieval methods based on conventional low-level feature descriptors and nature of the features with best set of features to select from. \cite{shao2013efficient} and \cite{ramezani2018retrieving} deal with employing Support Vector Machines (SVM) for video retrieval and relevance feedback mechanism for the retrieved results. 
\cite{liu2018research} use CNNs for feature extraction from video frames and PCA to reduce the feature dimension for retrieval of public cultural videos. \cite{petscharnig2018binary} leverage the power of CNNs and use the extracted features as frame descriptors with several similarity metrics for retrieving results for endoscopic medical database. \cite{zhang2019cnn} uses frame-level features extracted from CNNs with bag of visual words (BoVW) and clustering to retrieve videos based on visual content. The task of action recognition in video is closely related to retrieval as it also involves learning the content of the video based on visual information \cite{miech2017learning} employs discriminative clustering technique with a slightly modified version of Block-Coordinate Frank-Wolfe algorithm to recognize actions in videos. In \cite{miech2020end}, the same authors offer an end-to-end self supervised method to learn meaningful representations from videos for the tasks of action localization and segmentation. \cite{zolfaghari2018eco} offers an efficient way of using CNNs by capturing temporal dependencies among neighboring frames using late fusion that can be used on different video tasks.

Most of the works on video related tasks using deep learning was completely dominated by 3D convolutional neural network (3D CNNs) \cite{tran2015learning} and its variants with optical flow information. 3D CNNs by nature, use typically small 3D kernels to learn saptial and temporal features jointly by operating on spatial dimensions of the frame and temporal dimension which is across the frames. Though they perform well on standard tasks, they suffer from computational complexity, inability to learn specific attributes of objects in the videos as they learn spatio-temporal information cohesively which restricts them to be used in retrieval applications.

Later, \cite{donahue2015long} introduced LRCN, the usage of recurrent networks with CNNs for video caption generation task and J. Y. Ng et al. \cite{yue2015beyond} came up with an architecture to use 2D CNNs and RNNs with optical flow for video classification. The two aforementioned works proved the potential and effectiveness of coupling CNNs and RNNs together for video related tasks like classification and captioning. Later \cite{gu2016supervised} proposed a method using 2D CNNs and RNNs for video retrieval by calculating the embedding loss between two different input to the network. The variants of using 2D CNNs with RNNs can over come the computational complexity of being light weight in nature. RNNs are inherently capable of learning temporal sequences well and videos are constituted by actions of objects under focus. The motion in the videos are learnt well by RNN and CNNs are proven over the years to perform on image related tasks.

Though many methods addressed throughout the paper are able to operate with great accuracy of retrieval, they often suffer from inefficiency in terms of time of retrieval due to their heavy and complex nature which hinders them to be used in real-world, practical applications. Several works offer intuitive techniques for optimization and scalability of retrieval tasks such applying PCA for dimensionality reduction \cite{jegou2012negative} of feature representation for faster comparison, faster indexing in retrieval tasks \cite{sadeghi2019scalable}, applying different clustering methods \cite{ray1999determination} for easier matching, retrieval and approximation of user input queries etc \cite{suprem2018approximate}. These modular improvements can help in designing a retrieval framework which can work effectively in practical applications where precise and instantaneous extraction of results is of primary focus.

\section{Datasets}
The datasets that were used for this study are discussed in this section. For evaluating the capabilities of our proposed hybrid architecture and for the results of the discussed experiments, we use both image and video datasets.

\subsection{Image dataset}
For the experiment to analyse the effect of color on retrieval, we use CIFAR10 \cite{krizhevsky2009learning} which consists of 10 classes of labelled 6000 tiny color images of size 32x32 per class. CIFAR10 is a widely used dataset to benchmark deep learning models on various image related tasks. We also use Oxford5k Buildings Dataset \cite{4270197}, a buildings dataset which is popular for image retrieval tasks with 5062 images pertaining to 11 different landmarks with relevant text queries.

\subsection{Video datasets}
\textit{Since there are enough works on image retrieval in the field and it is implicit that our architecture can well handle images, we focus on the retrieval capacity of our method on videos.}

For bench-marking the effectiveness of our model in comparison with the conventionally use C3D for video retrieval, we use four different video datasets. KTH Recognition of Human Actions dataset KTH-RHA \cite{schuldt2004recognizing} in short, consists of 6 classes of human actions with about 150 grayscale videos per class. Human Motion DataBase 51 or HMDB51 \cite{kuehne2011hmdb} contains 7000 clips of human actions distributed across 51 class collected from various sources. UCF50 \cite{reddy2013recognizing} and UCF101 \cite{soomro2012ucf101} are action recognition datasets collected from youtube containing 50 and 101 classes respectively with around 100 clips per class.

To mimic a large video repository as seen in practical retrieval applications, the models were trained on UCF50 dataset and were tested on other 3 datasets containing about 4000 videos. This conglomerate of different datasets is essential as practical scenarios often involve a diverse range of visual input to the retrieval system demanding it to be reliable and robust. So, testing on organized, homogeneous data won't reflect the true ability of the solution to work in the real-world. 

To ensure consistency of results, it is made sure that there is no overlap between video classes in train and test sets. During the testing process, the videos are segmented into two main categories - seen and unseen. The seen category contains classes similar to which the model was trained on and the unseen category contains classes which are completely unrelated to the videos the model has been exposed to.

\section{Methodology}
We elaborate our proposed methods, their effectiveness, the intuition behind choosing them with the significance of employing them in practical applications.

\begin{figure}[htb]
    \begin{center}
        \includegraphics[width=8cm]{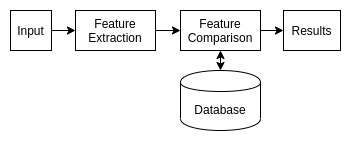}
        \caption{General Overview of CBIR}
        \label{Figure 1}
    \end{center}
\end{figure}

Content Based Media Retrieval (CBMR) can generally be defined as the computer vision technique of retrieving media that is relevant to the content of the input from a large database. In general, any content based retrieval task can be denoted by a simple block diagram as shown in Figure \ref{Figure 1}

The architecture we propose consists of the following novel properties:
\begin{enumerate}
    \item Our hybrid framework works for both images and videos in tandem.
    \item It is modular in nature hence enabling every part of it to be replaced by an appropriate alternative.
    \item Our architecture is light, fast and efficient, thus making it a scalable alternative over the other heavy architectures.
    \item Our new re-ranking method is simple yet scrupulous and could be extended to any retrieval task not restricted to images and videos.
\end{enumerate}

\subsection{Proposed architecture}
\begin{figure*}
  \begin{center}
    \includegraphics[width=12cm]{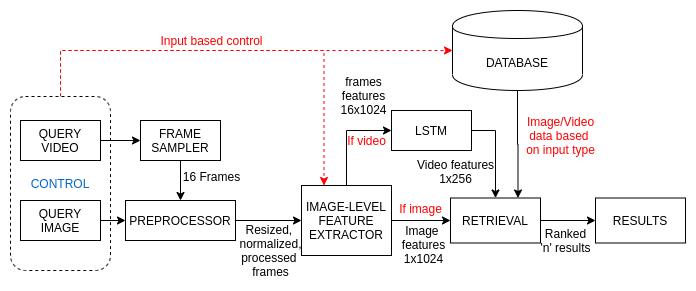}
    \caption{Proposed recurrent convolutional architecture (2D CNN + LSTM).}
    \label{Figure 2}
    \footnotesize{The text marked in red shows the conditional flow of data. As seen, an image is directly preprocesed, feature extracted and is sent to comparison. The video frames are sampled, preprocessed, feature extracted, processed through LSTM that generates final feature representation for comparison with other videos}
  \end{center}
\end{figure*}

We introduce our novel hybrid architecture for content based visual media retrieval that works on both images and videos conjointly in this section. This architecture consists of co-existing plug-and-play models that the user can readily select and deploy depending on the application and data. The proposed fused recurrent convolutional architecture which is inspired by 2D CNN + LSTM architecture \cite{hochreiter1997long} is described in Figure \ref{Figure 2}. 

A video is a collection of strongly correlated frames or images that are connected temporally and coherently holding information about an action. The architecture is primarily built around video operation as image is considered as a subset of a video. The architecture consists of an image feature extractor that can output the features based on the visual and abstract content in the input image. Any type of feature extractor -traditional extractors like SIFT, SURF, CENSURE etc. or CNN architectures like VGG16, MobileNet, ResNet, InceptionNet can be used depending on the nature and criticality of the application. We selected an ImageNet \cite{deng2009imagenet} pre-trained MobileNet model as the feature extractor because of its best performance. ImageNet \cite{deng2009imagenet} is by far the largest (1.2 million images) and most comprehensive image dataset covering a variety of images. This large corpus improves generalization in the feature extractor and since MobileNet V1 \cite{howard2017mobilenets} is known for its best performance with a small memory footprint which is important for fast operation. The output of the feature extractor is a feature embedding or representation of size \textit{d} that encapsulates the content of the input image. This feature representation is made use for comparison with that of the other indexed images (or videos) in the database in case the input of the retrieval system is an image and the retrieved results are ranked using our similarity metric and are rendered.

If the input of the retrieval system is a video, the video is first segmented into smaller chunks of clips of short duration. By heuristics, an average of 15 frames is sufficient to represent a video of 10 seconds duration at 30fps. Frame sampling rate is critical factor determining the performance of video retrieval system in terms of speed of operation and accuracy. Often, a sampling rate is selected so that a fine balance between both is achieved. Each of the chunks are then passed to a frame sampler which samples the frames at a rate as required by the LSTM \cite{hochreiter1997long} model. Our model operates on \textit{16 frames} per video. The sampled frames are pre-processed and passed to the aforementioned image feature extraction model. The set of feature representations \textit{(16 x d)} is passed to the LSTM model that is trained to learn the temporal attribution in videos in the similar fashion. The number of frames sampled (16)/ the number of \textit{time steps} in the LSTM model regulates unrolling with one frame feature at a LSTM time step.

LSTM \cite{hochreiter1997long} is capable of learning the temporal correlation between frames from the input image's latent representation and preservation of this information in the LSTM's cell state is vital to produce the best generalized video representation for retrieval purposes. The hidden state representation of the final time step after unrolling is taken as the video's final feature representation. This can be substituted by many other strategies like aggregating the hidden state at every time step as a feature representation, combining the cell state and hidden state from the final time step etc. depending on the performance of each configuration and the necessity for the application. The final representation is compared with the other indexed data from the repository using our efficient ranking algorithm based on which the retrieved results are rendered.

It can be easily seen that each and every module of the architecture is vital towards the overall performance. Also, a careful and strategical selection of the parameters can result in the best solution possible.

\subsection{Classifiers}
A classifier is a simple yet efficient solution to learn the content of the image and hence was chosen for our primary study. An image classifier has many attractive properties one of which is preserving generalized feature maps in the first few layers. MobileNet is a small yet efficient neural network architecture that can predict with high accuracy and speed. ImageNet contains around 1000 classes of images with objects which is required for the model generalization. Intuitively, an ImageNet trained model would have covered most of the common objects seen in videos. Transference of the learnt knowledge from models trained on large datasets can help in generalization since they preserve optimal weights from receptive fields learnt on a variety of images in a large corpus. This feature representation transfer can help in minimizing domain divergence and providing robust feature representations. After an input image is sent to the pretrained MobileNet, the activation of \textit{global average pooling} layer of dimension 1024 is used as image feature representation.

We use a single layer LSTM network as found it to be performing almost similar to having multi-layer LSTM. LSTM is also trained as a video classifier whose input is a set of 16 frame level features and a softmax probability distribution over the number of classes as its output. We snip the LSTM model to obtain the final hidden representation as the feature embedding for the video. This dimension depends on the number of LSTM units which is 256 in our case.

\begin{figure}
    \begin{center}
        \includegraphics[width=7cm]{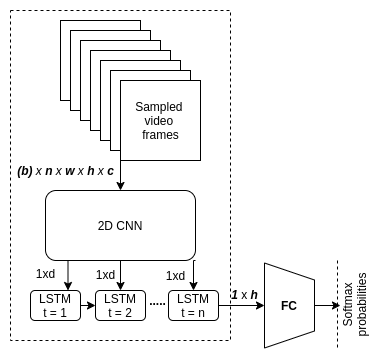}
        \caption{End to end training of the proposed architecture.}
        \footnotesize{\textbf{b} is batch size which we will ignore for now, \textbf{n} is the number of frames per video / LSTM time steps , \textbf{w,h} is the resolution of the frames and \textbf{c} is the number of channels. The dotted enclosure shows the part of the architecture that will be used for retrieval in the final stages.}
        \label{Figure 3}
    \end{center}
\end{figure}

Alternatively, if the retrieval process is on a confined data space for the application, the 2D CNN and the LSTM architecture enables end-to-end training as a classifier as shown in Figure \ref{Figure 3}. It should be noted that the end-to-end training works only for the data whose distribution or nature is known prior to training and it is similar to the 3D CNN in terms of operation and the generalization ability to learn the objects in the frames is low.

\subsection{Training configuration}
We train the 2D CNN + LSTM model and the 3D CNN model on the same dataset to benchmark the performance against each other. For training, we take a subset of UCF50 with 38 classes which have almost equal distribution of videos i.e. more than 120 videos per class. The dataset consists of 4557 videos and we use 90\% of it for training and 10\% of it for validation. For ensuring the best performance, we augment the data using random sampling and central cropping of videos.

We use almost the same hyperparameters for training both 2D CNN + LSTM and 3D CNN models. Adam optimizer is used as it is a better alternative over gradient descent for faster convergence and adaptive learning.  We use dropout of 25\% after the fully-connected layers in each of the networks as regularization helps in generalization, and ReLU activation function is used as it reduces training time. We use categorical cross-entropy loss to train the classifiers. The details of training\footnote{Models trained on Intel Xeon CPU, 12GB RAM, 16GB Tesla P100 GPU on inputs of size $16 \times 224 \times 224 \times 3$} are given in Table \ref{tab: Table A}.

\begin{table}
    \centering
    \caption{Training information of models}
    \label{tab: Table A}
    \begin{tabular}{|p{3cm}|c|c|}
        \hline
        Hyperparameter & 3D CNN & 2D CNN + LSTM \\
        \hline
        Epochs & 75 & 100 \\
        Batch Size & 4 & 64\\
        ADAM - Learning Rate & 0.005 & 0.001\\
        Feature representation dimension & 256 & 256\\
        \hline
    \end{tabular}
\end{table}

\subsection{Results ranking}
For retrieval tasks, euclidean distance and cosine similarity are the most used comparison metrics. Euclidean distance / L2 Norm calculates the distance between two points in the euclidean space and is given by the equation \ref{eq 1}
\begin{equation}
     \label{eq 1}
     ED\left( p,q\right)   = \sqrt {\sum _{i=1}^{n}  \left( q_{i}-p_{i}\right)^2 } 
\end{equation}
Cosine similarity measures the similarity between two vectors using inner product in multidimensional space. Cosine similarity is given by equation \ref{eq 2},
\begin{equation}
    \label{eq 2}
    \cos{\theta} = cos(p, q) = \frac {p \cdot q}{||p|| \cdot ||q||}
\end{equation}

The euclidean distance measures how two vectors are far apart and the cosine similarity measures the orientation between two vectors. Usually only one of the two metrics are used for retrieval until \cite{tanioka2019fast} used L2 norm to approximate cosine similarity for faster calculation. \cite{wang2010new} came up with a solution to integrate distance and rotation of features in multidimensional space to match and compare them.

Inspired by these two works, we devised a simple and novel ranking and filtering approach called the \textit{proximal affinity re-ranking} (PAR) by using a combination of euclidean distance / L2 Norm and cosine similarity based ranking method. The \textit{top k} results are first extracted based on euclidean distances from closest to farthest among the data in the repository. This process aggregates the closest feature representations in the latent space. We then threshold $\delta_t$ and short-list \textit{s} results based on the value of cosine similarity which selects the feature representations that have similar orientation from the origin in latent space. The algorithm for proximal affinity re-ranking is shown below in Algorithm \ref{Algorithm 1}.

The intuition behind this process is that similar representations are ideally concentrated in the same direction in parts of the latent space. To render the best results, the closer ones with similar orientations to that of the input data are selected as they represent the similarity in terms of content provided that the representations are robust and from a generalized model. The value of the threshold $\delta_t$ is chosen empirically depending on the nature of the model's output representations and performance on a sample test set. This threshold helps in filtering out irrelevant results among the retrieved ones.

\begin{algorithm}
\caption{\label{Algorithm 1}Proximal affinity re-ranking}

\SetAlgoLined
\KwIn{Query data $x_i$\\ \hspace{10mm} Feature extraction model $f$\\\hspace{10mm} Indexed features of n data samples in the repository $Y_{DR} = [y_1, y_2, .. y_n]$}

\KwOut{Indices of retrieved data samples, $R_{indices}$}
$y_i = f(x_i)$ \Comment{\hspace{17.5mm}\# \footnotesize{Feature representation of input}}\\ 
$E = []$ \Comment{\hspace{24mm}\# \footnotesize{Array of euclidean \textit{n} distances}}\\ 
\For{j = 1,2,..n}
    {
    $E[j] = ED(y_i, y_j)$
    }
$E_s = sort\_k(E, k)$ \Comment{\hspace{6.5mm}\# \footnotesize{Sorts and returns closest \textit{k} samples}}\\ 
$E_{si} = indices(E_s)$ \Comment{\hspace{6mm}\# \footnotesize{Returns indices of the values}}\\ 
$R_{indices}$ = [] \Comment{\hspace{15mm}\# \footnotesize{Indices of the retrieved samples}}\\ 
add\_count = 0\\
\For{l = 1,2,..k}
    {
    $c_l = cos(y_i, y_l)$\\
    \If{$c_l > \delta_t$}
        {
        $R_{indices}$[add\_count] = $E_{si}[l]$\\
        add\_count = add\_count + 1
        }
    }
\end{algorithm}

\section{Experiments}
We describe the experiments involved in our work in this section.
\subsection{Effect of color on retrieval}
Though, it is established that CNNs benefit from learning colors resulting in better performance on classification tasks, the exact dependence on color for retrieval tasks has not been studied in detail. We were curious on how much color in images can make a difference in terms of performance of retrieval in comparison to grayscale images. For this experiment, we shallow trained two instances of the same MobileNet V1 \cite{howard2017mobilenets} classifiers with exactly the same parameters except for only the number of channels – 3 for color and 1 for grayscale on the same train data and tested them on unseen data till the validation accuracy reaches about 70\%. The train (seen) and test (unseen) sets were selected randomly with each about half of the number of classes in respective datasets.

\subsection{Video Retrieval}

Our experiments on video retrieval was two-fold as shown below.
\begin{enumerate}
    \item Testing the efficiency of our proposed framework, 2D CNN + LSTM against the conventional 3D CNN.
    \item Checking the performance difference between the proposed proximal affinity re-ranking method and euclidean distance, the conventionally used metric.
\end{enumerate}
The models were trained and tested in the same environment. All the critical hyper-parameters were kept the same between the models to accurately benchmark the results against one another. Testing a retrieval system on data similar to the train data will reduce to a classification problem. The potential of the system can be measured only on unseen and irrelevant data. Also, to mimic the environment containing new or unseen videos which can be considered as samples out of the train data distribution, we have created a test environment in which there are two different sets of data - \textit{seen} and \textit{unseen}. Since we had multiple datasets for our utilization, we created a conglomerate of the test data but making sure that no video from the training dataset was reused in anyway for testing. The data in the test pool from different datasets was segregated into two sets - one with classes of video samples similar to the ones available in the train data as \textit{seen} and the \textit{unseen} set consisted of all the other data that were completely irrelevant to the data that was used for training to check the generalization of models and the robustness in representations. The unseen data resembles operating in the real world where one can't expect the models to only perform on the data they are familiar with. The basic code for our work is available at \href{https://github.com/ambareeshravi/media_retrieval}{https://github.com/ambareeshravi/media{\textunderscore}retrieval}

\section{Results}
The results of our experiment on the effect of color on retrieval performance is given in the Table \ref{tab:Table B} below. The MobileNet color and grayscale classification models were trained and tested on CIFAR10 after which the fully-connected layers responsible for classification were discarded to get the feature representations. Models trained for 50 epochs on 32x32 images with validation accuracy of ~74\% and Oxford 5K dataset with validation accuracy ~69\% for classification. It is apparent from the Table \ref{tab:Table B} that the difference in accuracy between the grayscale model and the color model on untrained classes is meagre. This permits the grayscale models to be used in retrieval tasks in which computation efficiency is critical as using grayscale inputs ensures faster computation as it has 4\% lesser number of parameters (in MobileNet V1) and performance almost similar to color images. This parameter reduction will aggregate to a much larger number for videos while operating at scale on huge data repositories.

\begin{table}[htb]
    \begin{center}
    \caption{\label{tab:Table B} Image retrieval accuracy of color and grayscale models}
    \begin{tabular}{|p {3.25cm}|p {2cm}|p {2cm}|}
    \hline
    \multirow{2}{*}{Dataset} & \multicolumn{2}{|c|}{Retrieval accuracy in \% - top 100} \\
    \cline{2-3}
     & Color model & Grayscale model \\
    \hline
    CIFAR10 (untrained classes) & 81.62 & 80.03 \\
    \hline
    \end{tabular}
    \end{center}
\end{table}

The retrieval performance of the proposed 2D CNN + LSTM framework was compared with the conventional 3D CNN architecture on seen and unseen data which were segregated in 90\% and 10\% proportion for test videos to be retrieved and input seed videos respectively. Each seed video had about 100 similar videos to be retrieved. The salient features of our proposed framework against the conventionally used 3D CNN model is shown in Table \ref{tab:Table C}.

\begin{table}
    \begin{center}
        \caption{\label{tab:Table C} Notable attributes of the proposed architecture}
        \begin{tabular}{|m{3.5cm}|m{1.5cm}|m{1.5cm}|}
            \hline
            Parameter & 2D CNN + LSTM [ours] & 3D CNN \\
            \hline
            Total training time on 4500 videos & 2.7 hours & 16 hours\\
            Total number of parameters & 4.5 Million & 46 Million\\
            Feature extraction time for a 16x224x224x3 video on CPU & 0.8 seconds & 1.2 seconds\\
            Total model size & 22.5MB & 305.17MB\\
            \hline
        \end{tabular}
    \end{center}
\end{table}

The metrics which are most commonly used for retrieval tasks are accuracy, precision, recall and F1 score which are given by the equations \ref{eq 3}, \ref{eq 4}, \ref{eq 5}, \ref{eq 6} respectively. The correctness of the retrieved result is decided based on the category of retrieval and the input seed video.

\begin{equation}
    \label{eq 3}
    Accuracy = \frac{True Positives + True Negatives}{Total}
\end{equation}
\begin{equation}
    \label{eq 4}
    Precision = \frac{True Positives}{True Positives + False Positives}
\end{equation}
\begin{equation}
    \label{eq 5}
    Recall = \frac{True Positives}{True Positives + False Negatives}
\end{equation}
\begin{equation}
    \label{eq 6}
    F1 Score = \frac{2 *Precision * Recall}{Precision +  Recall}
\end{equation}

\begin{center}
    \begin{figure*}
        \centering
        \includegraphics[width = 15.5cm]{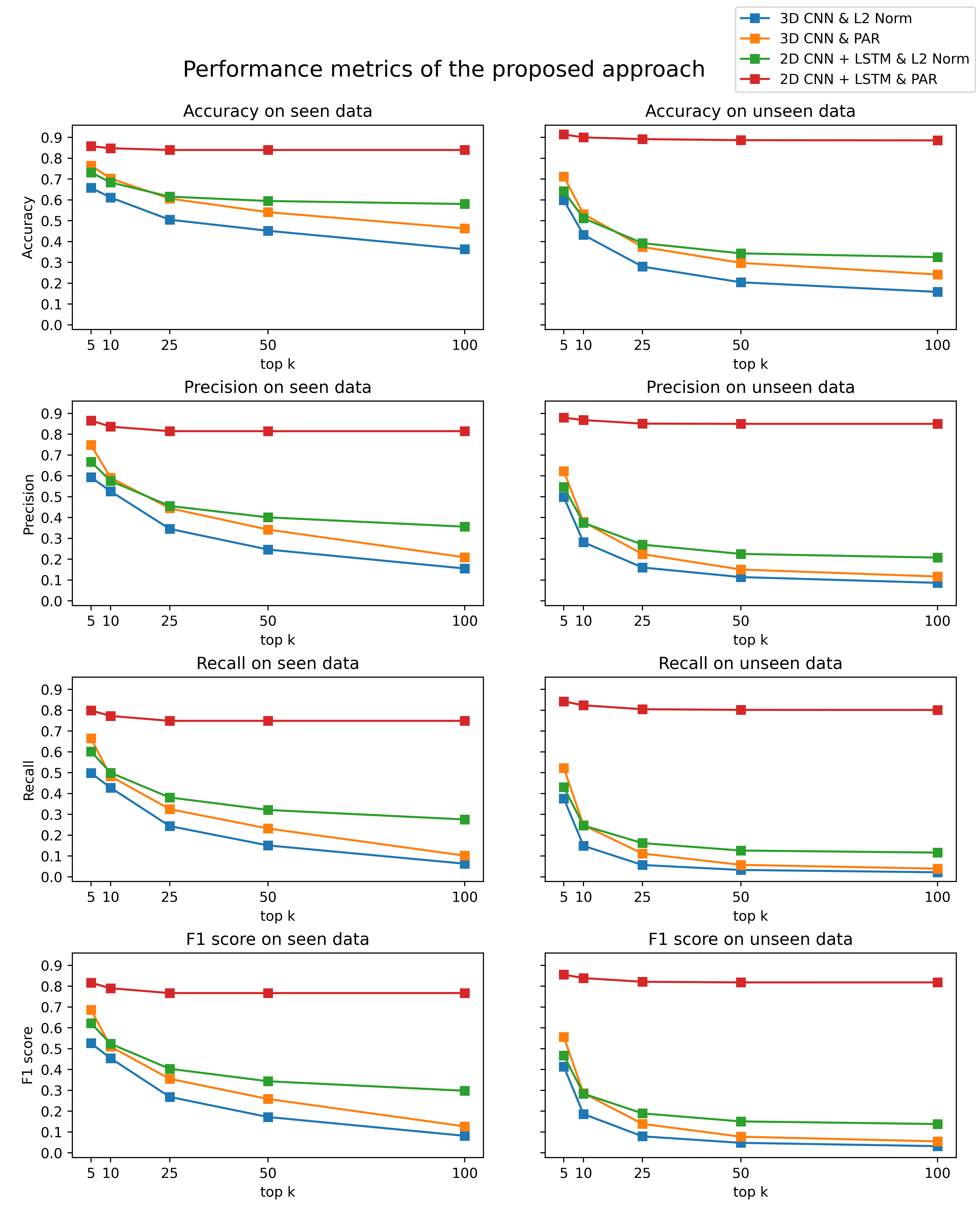}
        \caption{\label{fig: figure 4} Performance comparison of the Conv2D LSTM model with the prevalent method - trained on UCF50 and tested on KTH-RHA, UCF101 and HMDB50}
    \end{figure*}
\end{center}

The performance metrics for top 5 retrieved results are shown in Table \ref{tab:perf_comp} and the overall performance for the proposed 2D CNN + LSTM model with respect to the conventionally used 3D CNN model on both L2 Norm and the proposed proximal affinity re-ranking as metric are shown in the Figure \ref{fig: figure 4}. It can be seen that the proposed 2D CNN + LSTM consistently outperforms the conventional 3D CNN in all configurations indicating its ability to better generalize over the data. Also, our proposed re-ranking metric significantly boosts the performance of models as it filters out irrelevant results thereby choosing quality over quantity. The other performance metrics showing the effectiveness of the proposed approach \textbf{\textit{2D CNN + LSTM with PAR}} is shown in Fig \ref{fig: figure 4}. It can also be seen from Table \ref{tab:Table C} that our approach is lighter and faster when compared to 3D CNN which makes it the favorable option for video retrieval.

\begin{table*}
    \centering
    \caption{\label{tab:perf_comp} Top 5 performance comparison of the Conv2D LSTM model and 3D CNN model- trained on UCF50 and tested on KTH-RHA, UCF101 and HMDB50}
    \begin{tabular}{|p{2cm}|p{1cm}|p{1cm}|p{1cm}|p{1cm}|p{1cm}|p{1cm}|p{1cm}|p{1cm}|}
        \hline
        \multirow{3}{*}{Metric} & \multicolumn{4}{|c|}{Seen data} & \multicolumn{4}{|c|}{Unseen data}\\
        \cline{2-9}
         & \multicolumn{2}{|c|}{3D CNN} & \multicolumn{2}{|c|}{2D CNN + LSTM} & \multicolumn{2}{|c|}{3D CNN} & \multicolumn{2}{|c|}{2D CNN + LSTM}\\
         \cline{2-9}
         & EU & PAR & EU & PAR & EU & PAR & EU & PAR \\
         \hline
         Accuracy & 0.6578 & 0.7316 & 0.7644 & 0.8575 & 0.5983 & 0.6414 & 0.7113 & 0.914\\
         Precision & 0.5933 & 0.6675 & 0.7481 & 0.8656 & 0.4984 & 0.5462 & 0.6225 & 0.8793\\
         Recall & 0.4983 & 0.6015 & 0.6644 & 0.7984 & 0.3759 & 0.4303 & 0.5225 & 0.842\\
         F1-score & 0.5261 & 0.6222 & 0.6866 & 0.8165 & 0.4141 & 0.467 & 0.5566 & 0.8555\\
        \hline
    \end{tabular}
\end{table*}

\section{Best practices}
The number of frames has a critical role in determining the learning and in turn the performance of retrieval. We suggest using PCA  \cite{jegou2012negative} on the feature representations to reduce the dimension for faster comparison as we verified that there wasn't any significant drop in accuracy after reducing the feature dimension to a lower dimension with a greater than or equal to Proportion of Variance of 90\%. We also found that picking a decently trained model performing better than the best trained model which would have considerable bias towards the train data. Also, the right length of the feature representation is to be chosen while modelling which might impact computation and the performance.

There are several options to optimize the deep learning
models like filter pruning, replacing fully-connected layers with 1D convolution layers etc. For applications with speed of retrieval as primary goal, one obvious option is to use clustering on the media in the database and use only the centroid feature vectors for comparison. But if the application requires precision, clustering may not be a good option as it is approximate, doesn't work well on new data samples and since one has to update the centroid every time a data point is added.

\section{Future scope}
In this section, we identify the future research that can be carried out to potentially improve on our current work in terms of optimization and better implementation. First, we focus on better implementation, for better accuracy of retrieval the changes that can be imparted into a system. Due to the modular nature of our proposed architecture, an inclusion of an application specific feature extractor is feasible. It is important to replace the classifier with AutoEncoders, Siamese Networks and GANs and study their performance on retrieval tasks. Since RNN plays a vital role in our architecture, implementing other configurations of RNNs like GRU, bidirectional RNNs, multilayered/stacked RNNs instead of LSTMs and studying their effect on performance is necessary. Increasing the number of time steps of the RNN \cite{hochreiter1997long}, experimenting with the optimal number hidden units for the best results, implementing dynamic, bidirectional, multilayered LSTMs can also be done to potentially improve the performance. Due to the modular nature of our proposed architecture, using fast comparison libraries like FAIR’s FAISS \cite{johnson2019billion}, Spotify’s modification of approximate nearest neighbors ANNOY index \cite{li2016lin} will help in improving the processing time when it comes to large databases as they have the ability to use GPU for large scale computation to speed up the retrieval process.

\section{Conclusion}
Performance and computational constraints of the prevalent algorithms have been prohibiting their use in video retrieval tasks and to address the issue, we offer a novel, fused framework for conjoint image and video retrieval with a new simple re-ranking method called proximal affinity re-ranking. We also present substantial results to enunciate the effectiveness of our approach. The experimental results exhibit the overall ability of our approach in terms of accuracy and speed, over the widely used ones. This proposed architecture is flexible and can potentially be extended to other forms of retrieval like audio, document retrieval etc. We also discuss our experiments, best practices that can help in improving existing architectures and the future scope to augment our current research. 

\bibliographystyle{plain}
\bibliography{bibliography.bib}

\end{document}